\documentclass[journal]{IEEEtran}
\ifCLASSINFOpdf
   \usepackage[pdftex]{graphicx}
\else
\fi
\usepackage{amsmath}
\ifCLASSOPTIONcompsoc
  \usepackage[caption=false,font=normalsize,labelfont=sf,textfont=sf]{subfig}
\else
  \usepackage[caption=false,font=footnotesize]{subfig}
\begin{document}
%
\title{Content-based Video Retrieval in Traffic Videos using Latent Dirichlet Allocation Topic Model}
%
%
%

\author{Mohammad~Kianpisheh}

%
%

\markboth{Journal of \LaTeX\ Class Files,~Vol.~14, No.~8, August~2015}%
{Shell \MakeLowercase{\textit{et al.}}: Bare Demo of IEEEtran.cls for IEEE Journals}
%



\maketitle

\begin{abstract}
Content-based video retrieval is one of the most challenging tasks in surveillance systems. In this study, Latent Dirichlet Allocation (LDA) topic model is used to annotate surveillance videos in an unsupervised manner. In scene understanding methods, some of the learned patterns are ambiguous and represents a mixture of atomic actions. To address the ambiguity issue in the proposed method, feature vectors, and the primary model are processed to obtain a secondary model which describes the scene with primitive patterns that lack any ambiguity. Experiments show performance improvement in the retrieval task compared to other topic model based methods. In terms of false positive and true positive responses, the proposed method achieves at least 80\% and 124\% improvement respectively. Four search strategies are proposed, and users can define and search for a variety of activities using the proposed query formulation which is based on topic models. In addition, the lightweight database in our method occupies much fewer storage which in turn speeds up the search procedure compared to the methods which are based on low-level features.
\end{abstract}

\begin{IEEEkeywords}
Content-based retrieval, Surveillance, Latent Dirichlet Allocation, Topic Model, Search strategies, Query formulation.
\end{IEEEkeywords}

%
\IEEEpeerreviewmaketitle
\section{Introduction}
\label{intro}  
\begin{figure}[t]
\centering
\includegraphics[width=0.5\textwidth]{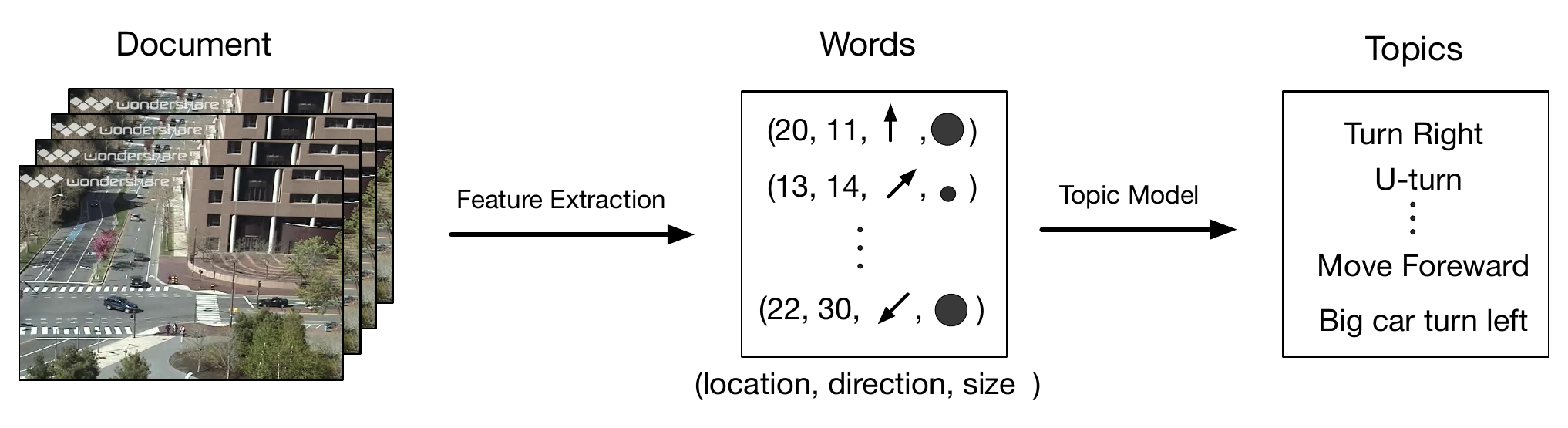}
\caption{Low-level features to high-level patterns procedure. First, visual features are extracted from each document which is a bag of visual features. Then, documents are fed into the LDA model, to discover semantically meaningful patterns (topics) in the scene.}
\label{fig:vidInput}
\end{figure}

\begin{figure}[t]
\centering
\includegraphics[width=1.8in]{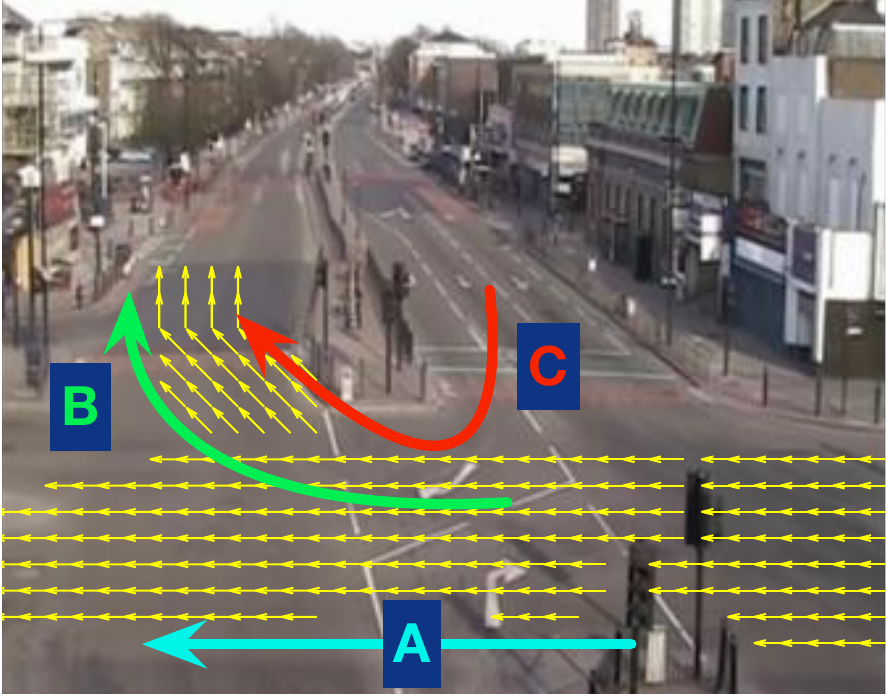}
\caption{An ambiguous topic including multiple minor actions.}
\label{fig:ambiguity}
\end{figure}

%
%
%
%
\IEEEPARstart{S}{urveillance} cameras are widely used in almost all public places such as traffic junctions, airports, subways, and shopping centers. Content-based retrieval is one of the most challenging tasks in surveillance videos, and it is highly influenced by the quality of video annotation. The massive amounts of video recorded daily makes it almost impossible to annotate them manually. Thus, unsupervised algorithms are needed to exploit semantically meaningful activities in video. \\ 
\indent Topic models are a class of hierarchical models, were originally proposed to discover meaningful topics in large text data collections in an unsupervised fashion. In these models, every document is considered as a mixture of topics, which in turn are a mixture of words. Processing the documents, the words that often co-exist in the same document are clustered into the same topic. After text domain promising results, topic models have been applied to a variety of data types such as images \cite{ref_12,ref_13}, sound signals \cite{ref_11} and surveillance videos \cite{ref_7,ref_9,ref_10}. In video domain, the input video sequence is segmented to short clips with 10 to 100 frames. Each clip is considered as a document and extracted visual features correspond to words, and by mining the documents, topics in the video sequence are exploited. In this case, a topic corresponds to an activity in the scene. The procedure from low-level visual features to high-level activities is illustrated in Fig.~\ref{fig:vidInput}. \\ 
%
\indent The performance of the retrieval system is highly affected by the accuracy of description space. For example, Fig.~\ref{fig:ambiguity} shows a sample topic in a junction learned using the Latent Dirichlet Allocation (LDA) topic model \cite{ref_5}. This topic contains multiple actions (A, B, and C), thus, the occurrence of each of these actions excites the illustrated topic in Fig.~\ref{fig:ambiguity}. In this case, it is impossible to infer which one of the minor actions (A, B, and C) has occurred in the respective clip.
%
The way of query definition and search procedure are other factors that influence the performance of the scene understanding systems. In this work, several techniques are proposed to address the ambiguity issue explained above. In addition, four search strategies are proposed which allow users to accurately define and search for different queries using the proposed query formulation. The rest of this paper is organized as follows: Section~\ref{subsec:relatedWorks} provides an overview of the recent literature. The description of the proposed method is presented in Section~\ref{sec_1-2}. Experimental results for the proposed search strategies are presented in Section~\ref{sec:experiments}. Finally, Section~\ref{sec:conclusion} concludes the paper. 
%
%
\section{Related Work}
\label{subsec:relatedWorks}
\indent Most of the proposed works for scene understanding can be broadly grouped into two categories. Under the first category, objects in the scene are first detected and then tracked to obtain their trajectories in the scene  \cite{ref_17, ref_18}. Object trajectories are trained and used for scene understanding analysis. For example, motions are learned using clustering in \cite{ref_17}, where object trajectories are hierarchically clustered using spectral clustering, and in the second stage, each group of trajectories is further clustered into subcategories using the temporal information. Tracking based approaches are prone to tracking errors. In these approaches, it is likely to lose an object for several frames, especially in the crowded scenes, which results in an incomplete trajectory for that object \cite{ref_7}.\\ 
%
\indent Under the second category, visual features are extracted from video sequence without any object detection and tracking. In most of these pixel-based approaches, learning algorithms are used to index the video stream by high-level semantically meaningful patterns. There are several methods that are based on topic model which model the scene with high-level patterns \cite{ref_7,ref_9,ref_15,ref_16}. These methods succeed in indexing videos but none of them address the retrieval task in surveillance videos directly except for the work presented in \cite{ref_7}, which is based on Hierarchical Dirichlet Process (HDP) topic model \cite{ref_21}. In their work, the video stream is decomposed into clips each of them is annotated with a distribution over the learned topics. To search for an activity, they define the user query as a distribution over the learned topics. The clips within the database are then compared with the query distribution using the Kullback-Leibler divergence to find the clips that involve the user query. \\
%
\indent  Among pixel-based approaches, in method presented in the video is indexed with low-level visual features, unlike the scene understanding approaches in which the video is indexed with high-level patterns. They use Locality-sensitive hashing (LSH) to hash the extracted low-level features into a lightweight lookup table. They define the user query as a set of action components each of them composed of a set of structured low-level features called tree. After defining the query and the region of interest (ROI), a set of trees is assigned to each action component and the full responses are found using dynamic programming. 
%
\begin{figure}[!t]
  \centering
  \includegraphics[width=0.43\textwidth]{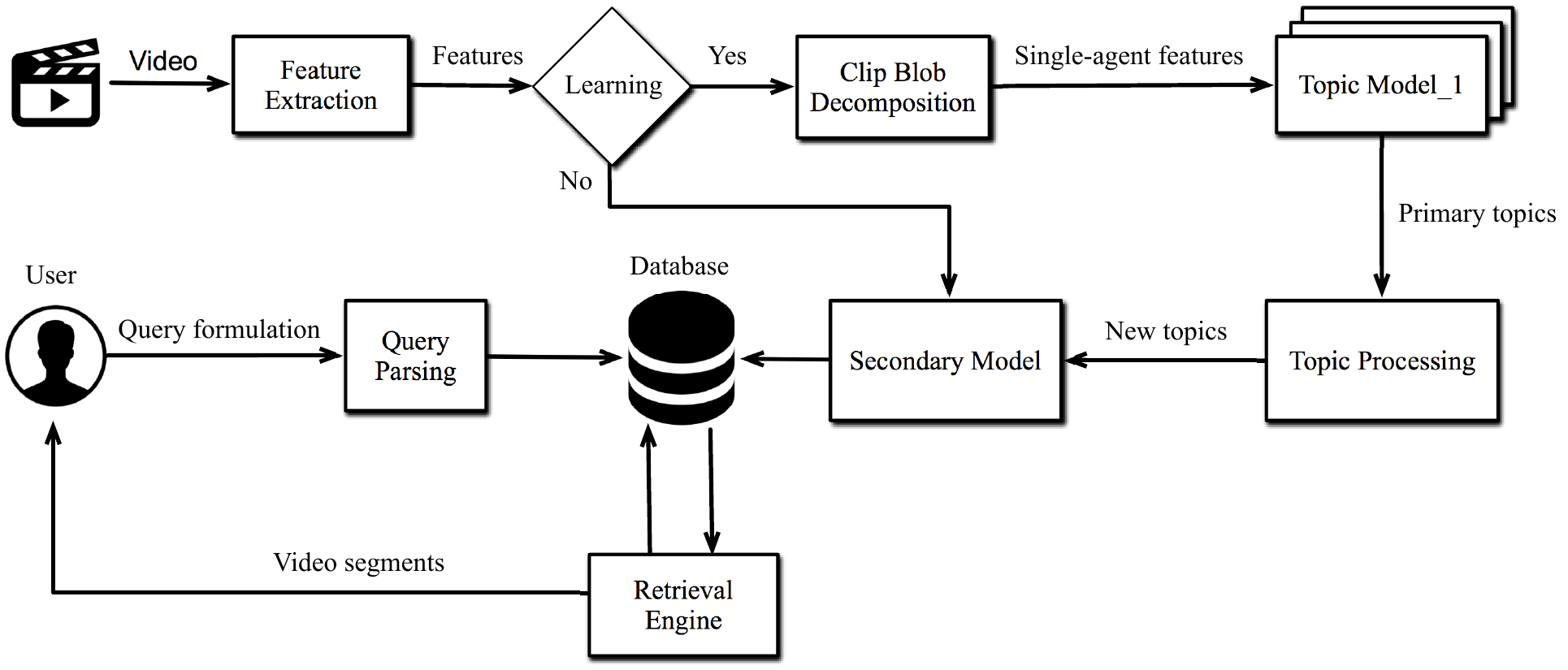}
\caption{The proposed method framework. In the learning phase, blob decomposition is performed on extracted features, and new feature vectors are fed into the corresponding LDA topic model. Topics of the primary model are processed to form the secondary model with primitive topics.  The user query formulation is parsed into the internal representation of the system, and the search procedure is performed using the proposed search strategies.}
\label{fig:framework}
\end{figure}
\section{Proposed method}  
\label{sec_1-2}
%
The framework of the proposed method is illustrated in Fig.~\ref{fig:framework}. As video streams in, low-level features: motion, persistence, and size are extracted. In the Clip Blob Decomposition step, Connected Component Labeling (CCL) algorithm is used to decompose each feature vector into several feature vectors each of them including a single-agent activity. The resulted feature vectors indicate activities that contain a single connected component. This pre-processing step increases the probability of obtaining topics which represent an activity consisting of a single blob. After the Clip Blob Decomposition step, in the learning phase, the resulted single-agent vectors are fed into the LDA topic model to discover activities in the scene. For each visual feature, a separate LDA model is learned to prevent multiple activities from different feature spaces to merge together into a single topic. In the Topic Processing step, the primary learned model is then processed to create a secondary model in which topics lack any ambiguity and each of them indicates a primitive action. In this step, first, each topic is decomposed into several topics using the CCL algorithm. By doing so, each topic in all feature spaces includes an activity comprising a single blob. In addition, motion topics are broken down into topics which consist of activities including motion in just one direction which results in even more primitive topics. After the learning phase, the secondary model is used to index the input video and creating the database. \\ 
\indent In the user side, the user can define his query using the proposed query formulation which is based on topic models, and specifies the way that user expresses his query. The Query Parsing specifies the way that the system parses the user query into the internal representation of the system, which in our case are topics. Having the user query determined, the Retrieval Engine searches the database and returns the video segments containing the queried action. Although primitive topics are void of ambiguity, these topics can not individually specify a complicated activity. To address this issue, in the proposed method a complicated activity can be defined as a sequence of primitive topics. Then, the desired sequence examples in the database are found using the Smith-Waterman \cite{ref_6} dynamic programming algorithm like the one presented in~\cite{ref_2}. In addition to topic sequence, three other search strategies including single topic, topic co-occurrence, and similar clips are considered in the Retrieval Engine to complete the system and allow the user to search for a variety of queries. \\  
%
\indent Indexing video with high-level patterns in the  proposed method results in a database with significantly lower size compared to the methods which are based on low-level features. A lightweight database, not only reduces the needed storage, but it also results in a significant speed up in the search procedure. The feature extraction step is also accelerated by leveraging the computing power of Graphics Processing Unit (GPU). 
%
\subsection{Low-level visual features}
\label{sec:FE}
In this work, the video sequence is uniformly segmented into non-overlapping clips each of them includes \textit{F} frames. Each frame within a clip is divided into $C\times C$ pixel-cells and the visual features including location, motion, persistence and size are computed for each cell. $F$ and $C$ are chosen regard to the frame rate, resolution, objects distance from the camera, and the scene type (e.g. junction, subway, etc.). \smallskip \\ 
%
\textbf{Location}: Most of the activities in the surveillance videos are characterize by the place they occur. Therefore, the location is considered in our feature extraction step. For a video with a frame size of $480\times 720$ which is divided into $10\times 10$ cells, we have $48\times 72$ pixel-cells in each frame. \smallskip \\
%
\textbf{Motion}: TVL1 optical flow method \cite{ref_1} is used to compute the motion features. Optical-flows greater than the $Th_{op}$ threshold are reliable and quantized into one of the eight cardinal directions, and areas in the scene where optical flow magnitude is lower than the $Th_{op}$ are considered as static. Finally, the most common direction in each pixel-cell is considered as the motion feature of that pixel-cell. \smallskip \\
%
\textbf{Persistence}: Persistence feature allows to detect static foreground objects in the scene, and more variety of activities can be captured using this visual feature. Persistence is occurring in areas in which an object belonging to the foreground, stays for a while in the scene and does not move. Thus, computing persistence needs background subtraction. Robustness to gradual and sudden changes in illumination is the main challenge in background subtraction methods in surveillance videos \cite{ref_3}. For example, trees shaken by the wind in the scene result sudden changes in the background pixel values. To cope with these issues, the Gaussian Mixture Model (GMM) background subtraction is used \cite{ref_4}. Modeling each pixel with a mixture of \textit{K} Gaussians which update continuously, makes GMM method robust to sudden and gradual changes in illumination. \smallskip \\
%
\textbf{Size}: After background subtraction step, blobs in the foreground can be detected using blob extraction, and the foreground objects can be further characterized. The size of each blob is considered as the number of pixels forming the blob, and blobs are categorized into two classes (small and large). Each pixel gets the size label (small and large) of the blob it belongs to, and each pixel-cell gets the most common size label of pixels that form it.    
%
\subsection{Clip blob decomposition (pre-processing)}
\label{subsect:clipBlobDecomp}
\begin{figure}[!t]
\centering
\subfloat[]{\includegraphics[width=1.6in]{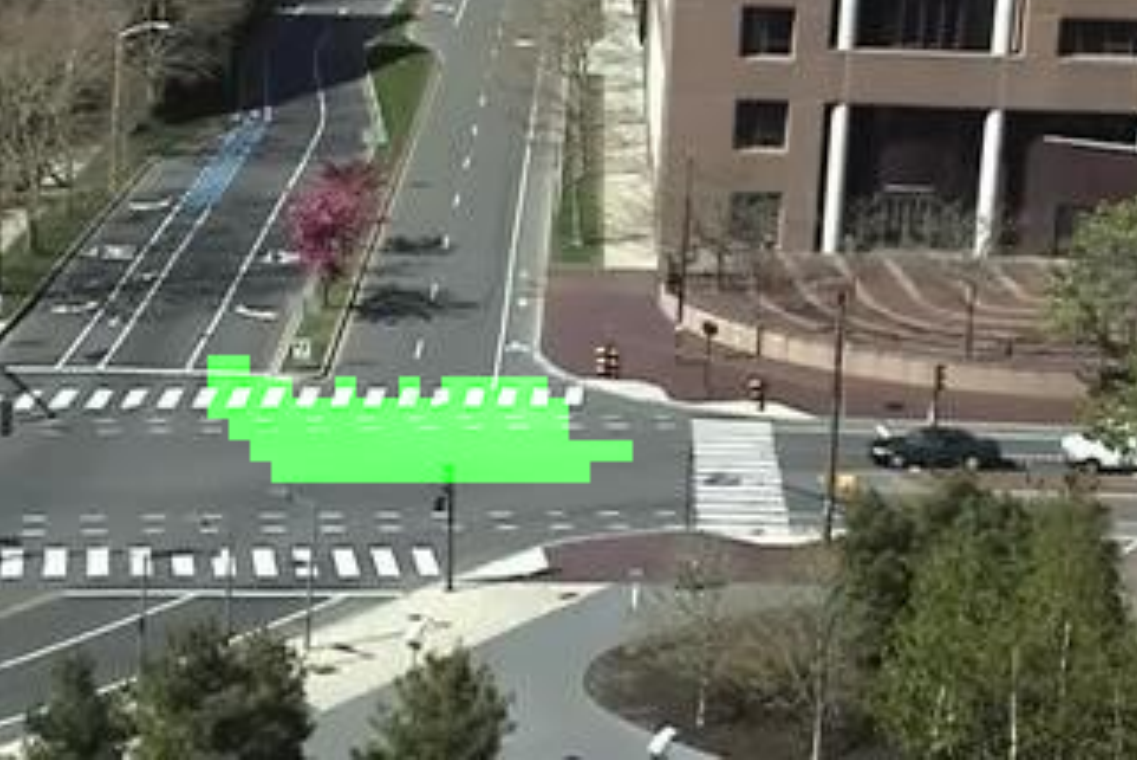}
\label{fig:docBlobDecomp1}}
\\
\subfloat[]{\includegraphics[width=1.6in]{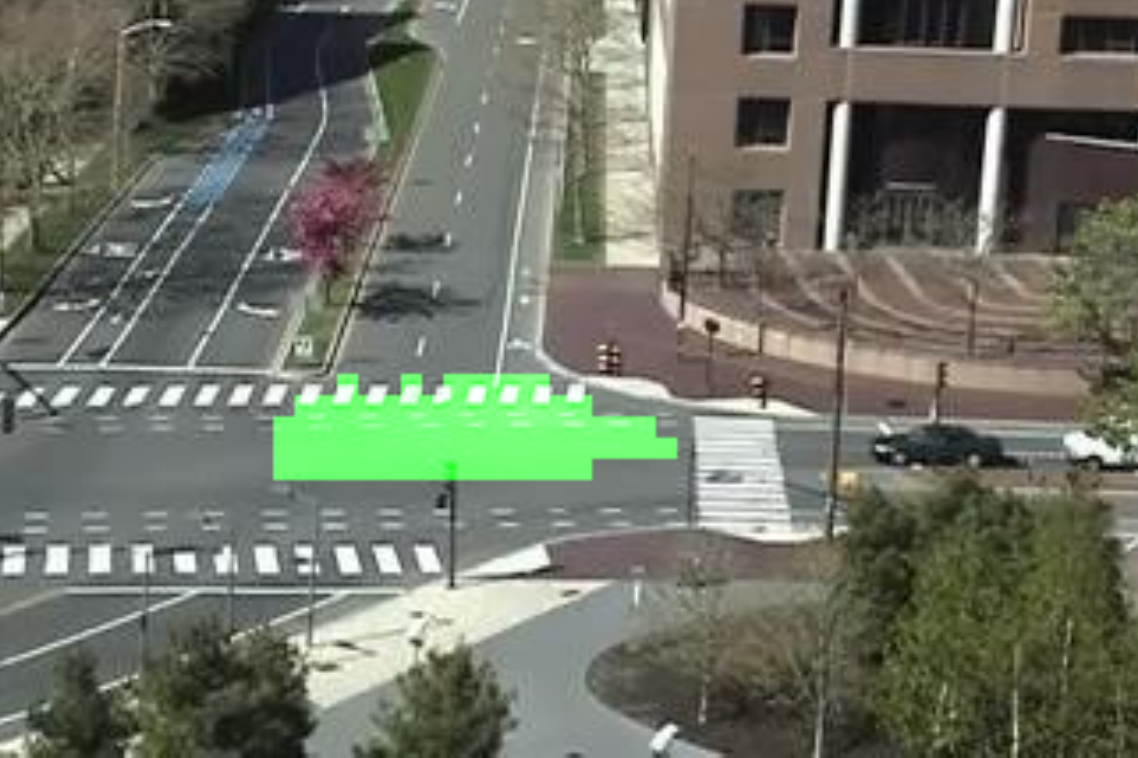}
\label{fig:docBlobDecomp2}}
\hfil
\subfloat[]{\includegraphics[width=1.6in]{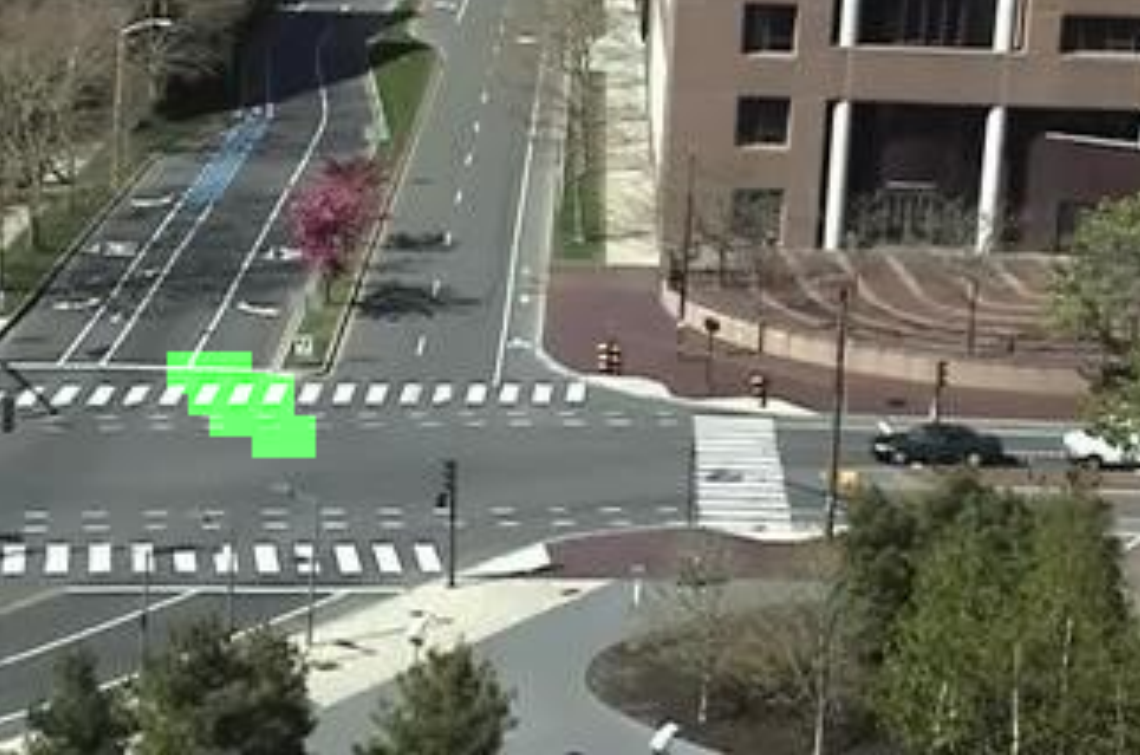}
\label{fig:docBlobDecomp3}}
\caption{Learned persistence topics: (a) a topic obtained without performing clip blob decomposition step, (b) and (c) spatially separate persistence topics obtained in the presence of clip blob decomposition step.}
\label{fig:docBlobDecomp}
\end{figure}

During the learning phase, the co-occurrence of multiple activities in the same clip may lead to a learned topic which represents two or more activities. In the proposed method, to prevent activities from mixing together, a pre-processing step is used before applying the LDA algorithm. By performing blob decomposition on each clip using the CCL algorithm, each feature vector is decomposed into new feature vectors each of them represents an activity consisting of a single blob which in turn reduces the probability of such merging process in the learning phase. Fig.~\ref{fig:docBlobDecomp}a shows a sample persistence topic obtained in a junction without performing the Clip Blob Decomposition pre-processing step (highlighted areas indicate persistence). Fig.~\ref{fig:docBlobDecomp}b and Fig. \ref{fig:docBlobDecomp}c show two persistence topics in the same area indicating two activities separated due to performing the Clip Blob Decomposition in our method.
%
%
\subsection{Separate models for each visual feature}
\label{subsec:sepModel}
Fig.~\ref{fig:qmulTopics} shows several learned topics using the LDA model in a Junction. Areas covered by dots in the figure indicate persistence. In the first experiment, the feature vector of each clip is made up of the combination of motion and persistence visual features. In this case, as shown in the first row of Fig.~\ref{fig:qmulTopics}, topics include activities from both feature spaces. These topics suffer from ambiguity issue in the feature space level, so they are not appropriate for the retrieval task. \\
%
\indent To get around this issue, a better approach is to learn a separate model for each visual feature. This approach not only prevents activities from different feature spaces to assign to the same topic, but it also reduces the number of visual words. For instance, consider a scene with $2400$ cells for location feature, eight main directions for motion feature, and two classes (small and large) for size feature. This setting, in the case of the compound feature vector, leads to a total of $2400\times 8\times 2=38400$ visual words, while the second approach leads to a motion feature vector with $2400\times 8=19200$ bins and a size feature vector with $2400$ bins. Thus, the number of words in the second approach is almost half of the first one. Some of the learned topics by separate models for motion and persistence features are shown in the second and third rows of Fig. \ref{fig:qmulTopics}. 
%
\subsection{Topic blob decomposition}
\label{subsect:topicBlobDecomp}
\begin{figure*}[!t] 
\centering
\subfloat[]{\includegraphics[width=1.7in]{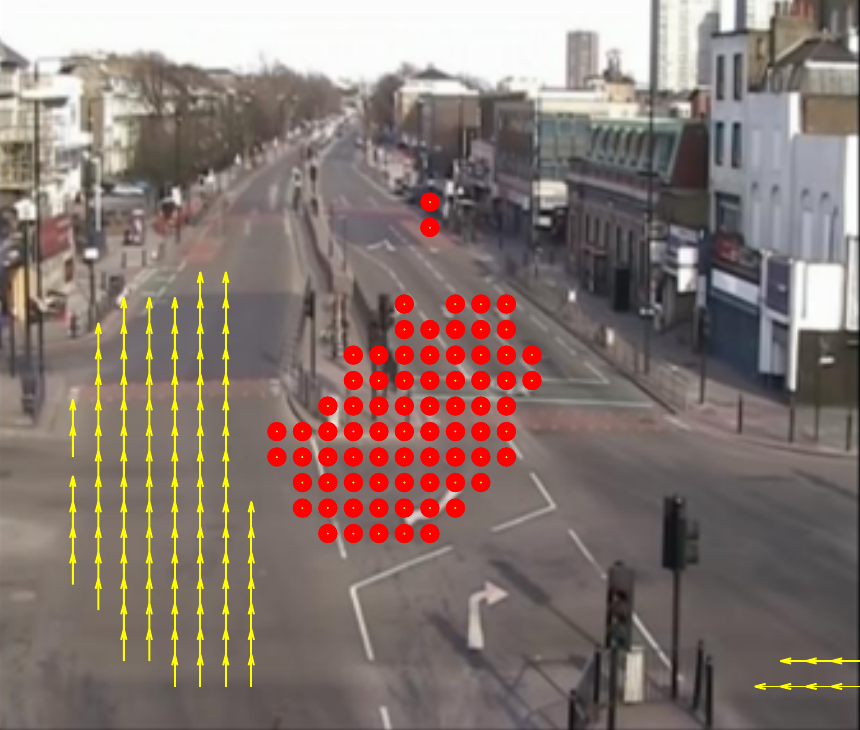}
\label{fig:qmulTopics1}}
\hspace{1em}
\subfloat[]{\includegraphics[width=1.7in]{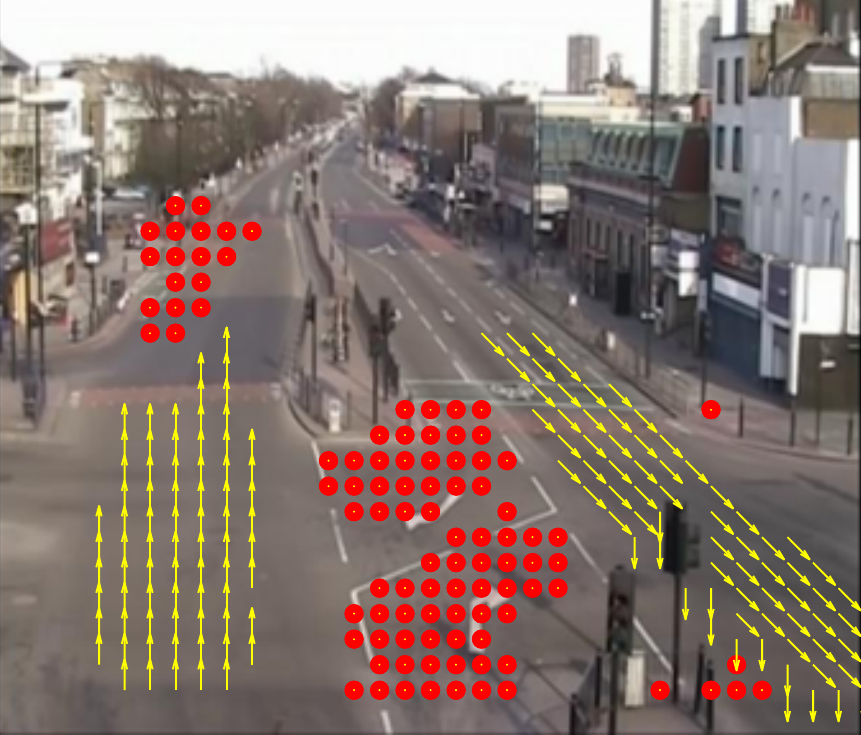}
\label{fig:qmulTopics2}}
\hspace{1em}
\subfloat[]{\includegraphics[width=1.7in]{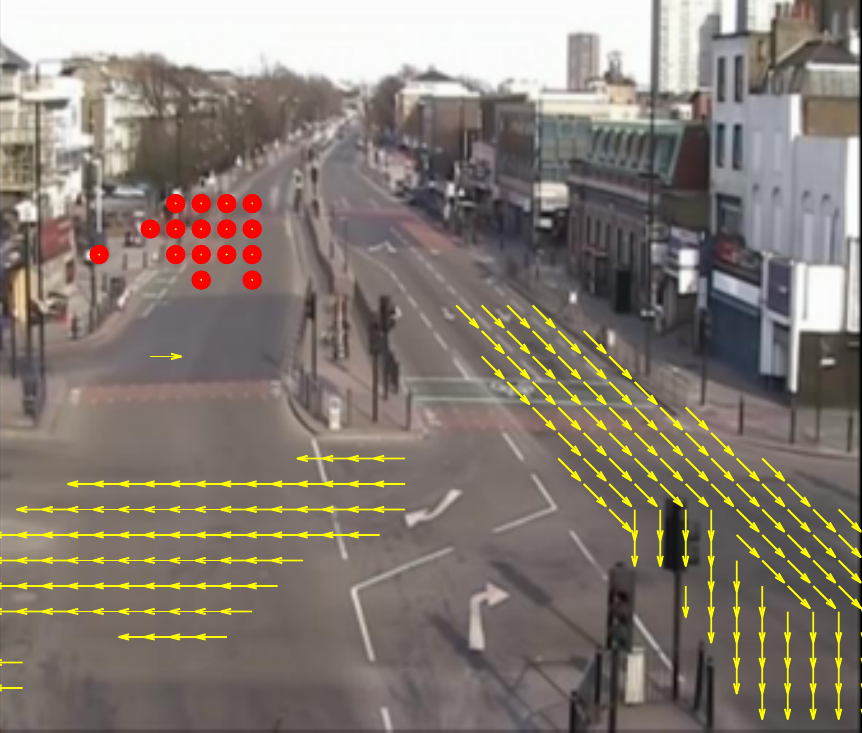}
\label{fig:qmulTopics3}}
\hfil
\subfloat[]{\includegraphics[width=1.7in]{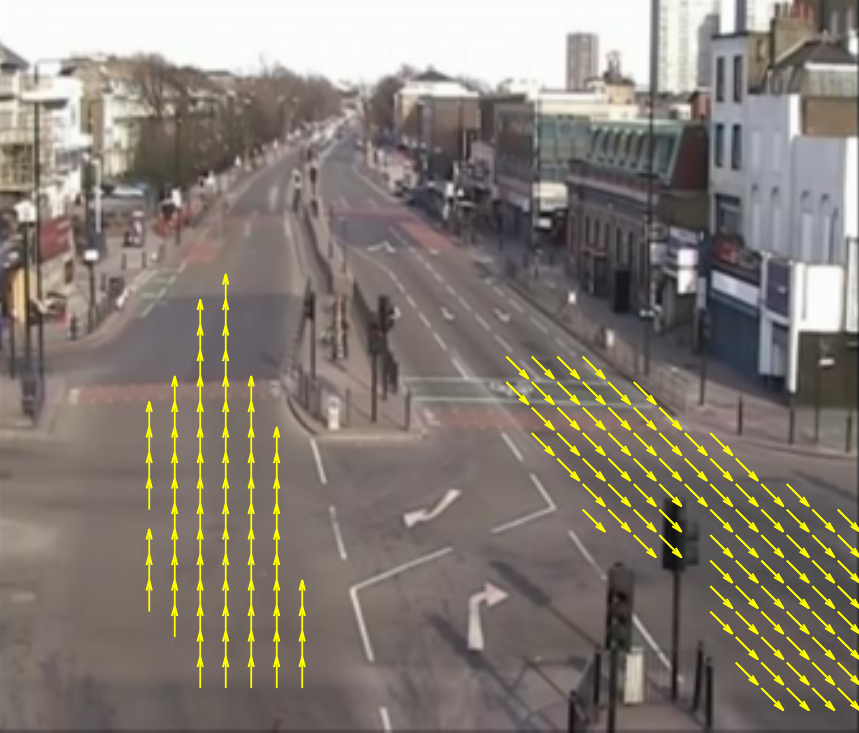}
\label{fig:qmulTopics4}}
\hspace{1em}
\subfloat[]{\includegraphics[width=1.7in]{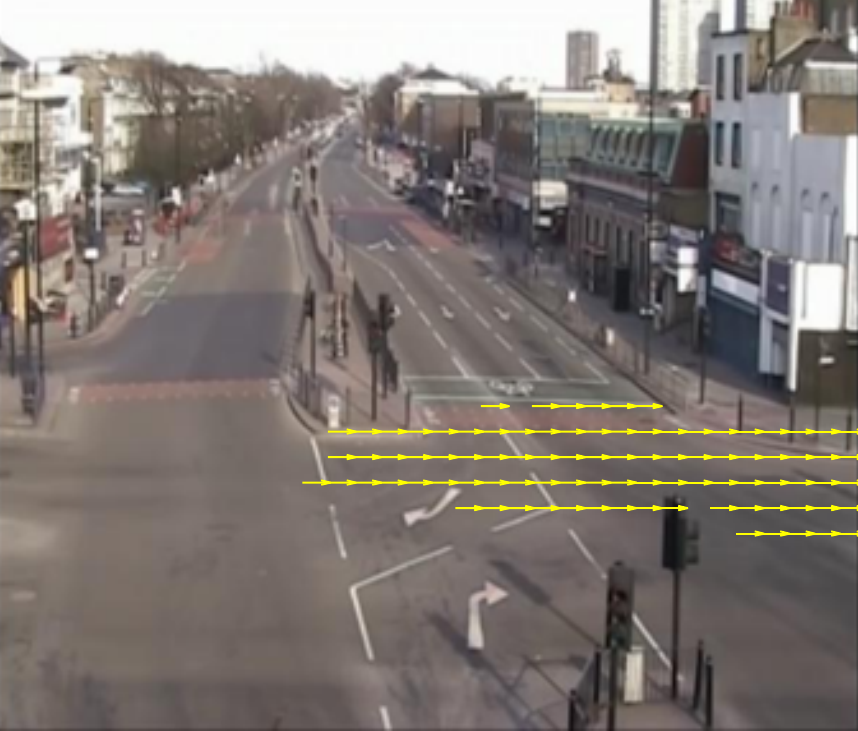}
\label{fig:qmulTopics5}}
\hspace{1em}
\subfloat[]{\includegraphics[width=1.7in]{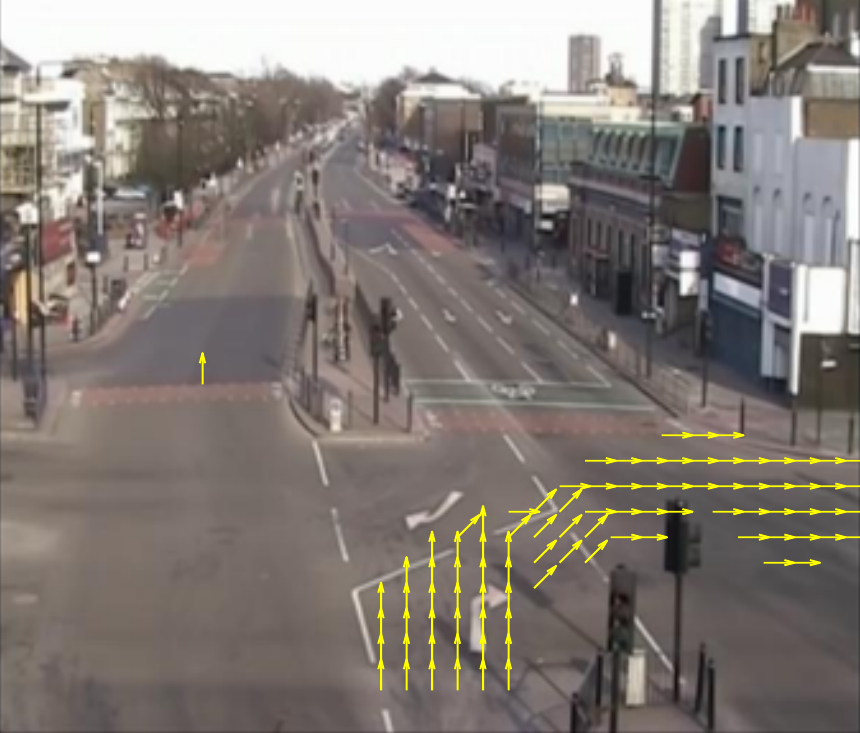}
\label{fig:qmulTopics6}}
\hfil
\subfloat[]{\includegraphics[width=1.7in]{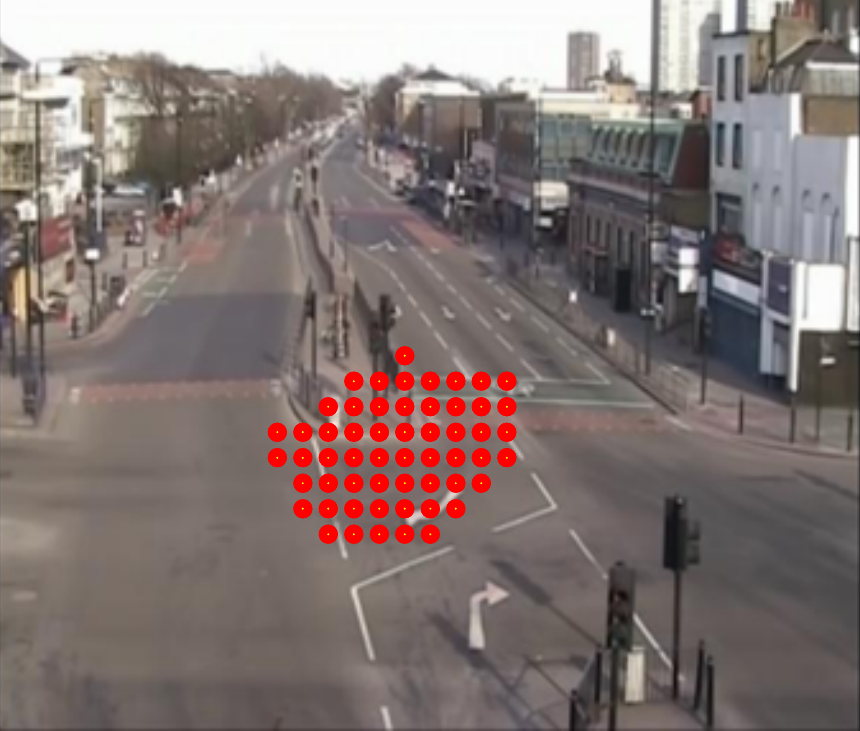}
\label{fig:qmulTopics7}}
\hspace{1em}
\subfloat[]{\includegraphics[width=1.7in]{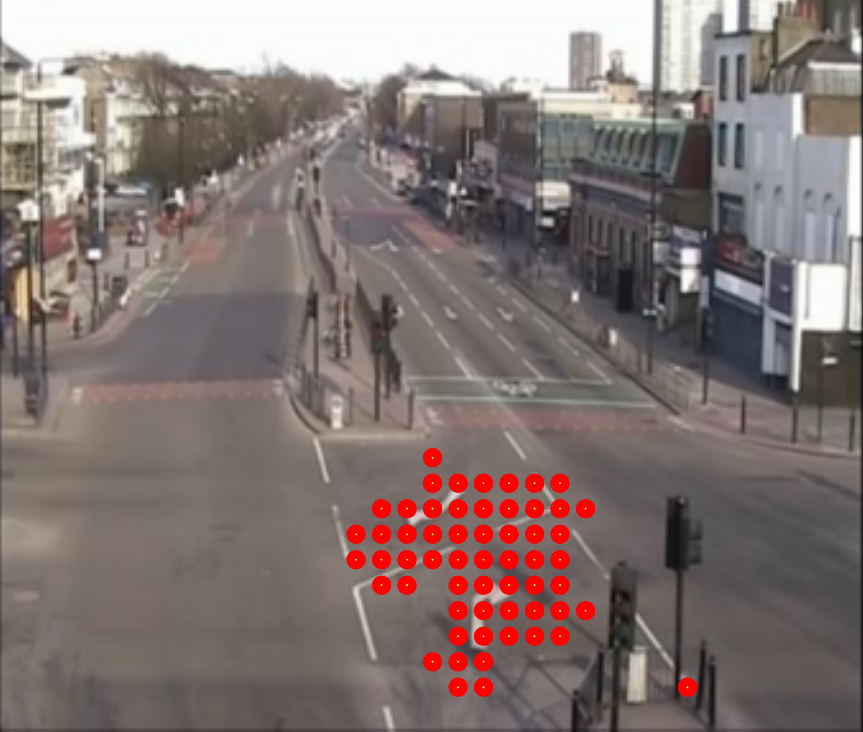}
\label{fig:qmulTopics8}}
\hspace{1em}
\subfloat[]{\includegraphics[width=1.7in]{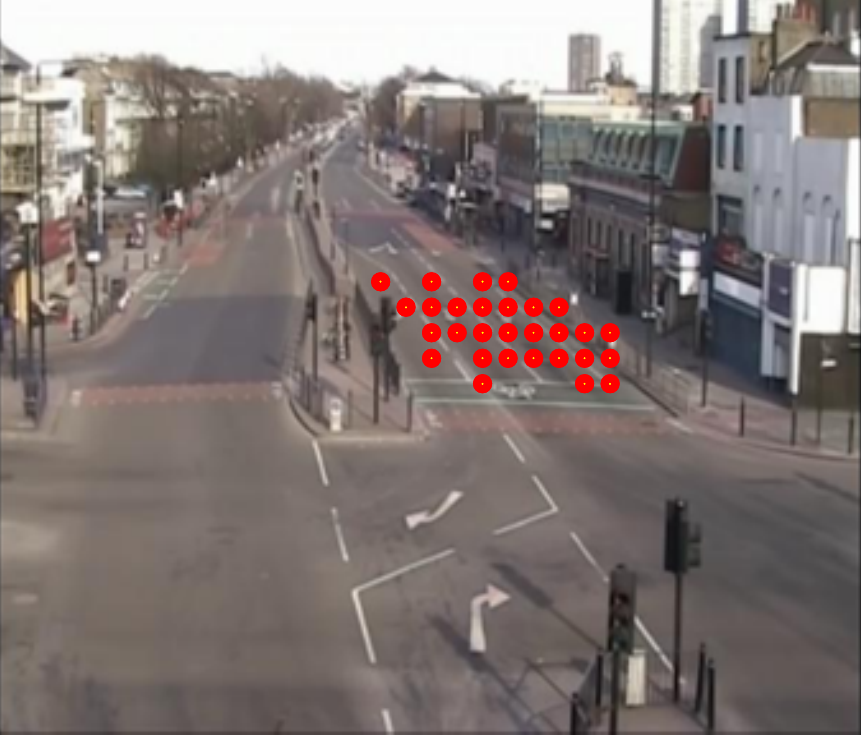}
\label{fig:qmulTopics9}}
\caption{ Sample topics learned using LDA model for (first row) the compound feature vector, (second row) the motion visual feature, and (third row) the persistence visual feature.}
\label{fig:qmulTopics}
\end{figure*}

Learning separate models for each visual feature does not obviate the ambiguity issue entirely. As can be seen from Fig.~\ref{fig:qmulTopics}d, a topic in a given feature space may include multiple activities. In Topic Processing step in the proposed method, this issue is simply solved by decomposing topics of the primary learned model into topics which each of them indicates a simpler activity. This procedure is done by blob decomposition of the learned topics using the CCL algorithm. Indeed, we create a new model in which each topic includes a single-agent activity. The secondary model is created by processing the topics of the primary learned model.
%
%
\subsection{Topic direction decomposition}
\label{subsec:dirDecomp}
\begin{figure}[!t]
\centering
\subfloat[]{\includegraphics[width=1.6in]{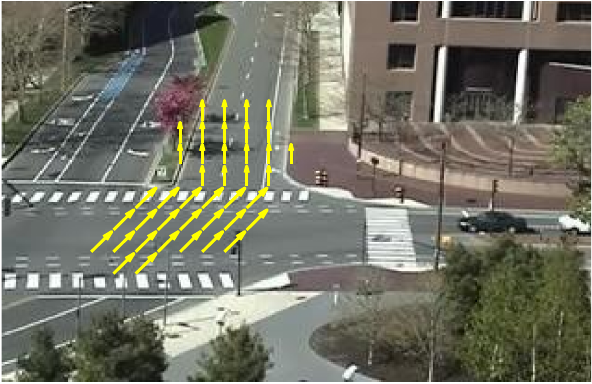}
\label{fig:dirDecomp1}} \\
\subfloat[]{\includegraphics[width=1.6in]{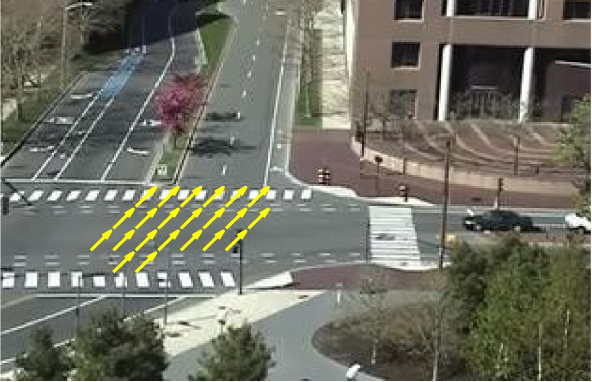}
\label{fig:dirDecomp2}}
\hfil
\subfloat[]{\includegraphics[width=1.6in]{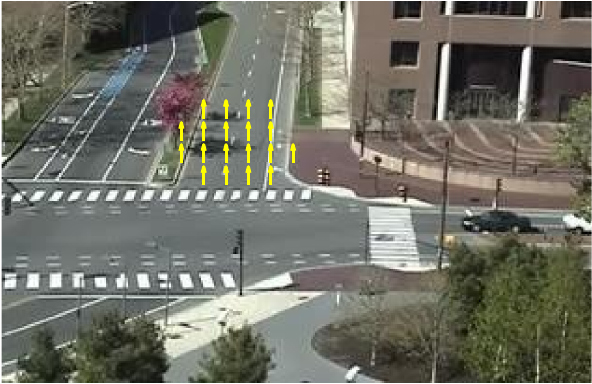}
\label{fig:dirDecomp3}}
\caption{Topic direction decomposition. (a) A sample motion topic containing movement in two directions, (b), (c) Primitive motion topics consisting of a single motion direction}
\label{fig:dirDecomp}
\end{figure}

To obtain even more primitive topics, after the blob decomposition step, the resulted motion topics are decomposed into topics which include motion in just a single direction. Fig~\ref{fig:dirDecomp} shows an example of topic motion direction decomposition. For better visualization, arrows are drawn for a bigger cell size. Topics including words in a single direction represent more primitive actions. 
%
Another issue regarding topics with multiple directions is that these topics may include two words with different directions in the same pixel-cell. In this case, topics may include two overlapping patterns. This is another form of ambiguity in topics which leads to a lot of false alarm responses.
%
After performing the blob and direction decomposition on topics, highly correlated topics can be removed to prevent redundant data storage. Note that correlated topics can be removed only if topics are primitive and represent simple actions. Indeed, two complicated topics may have considerable overlap but indicating two essentially different activities.
\subsection{Retrieval engine}
\label{subsec:tpSeqDP}
Adopting topic processing techniques results in a secondary model with primitive topics which are void of ambiguity. Although primitive topics lack any ambiguity, they can not be used individually to search for a complicated activity. To cope with this issue, in the proposed method, a complicated activity is defined as a sequence of primitive topics. The sequence length is determined by the user. These topic sequences are then searched in the database using the Smith-Waterman \cite{ref_6} dynamic programming like the one presented in \cite{ref_2}. In addition to topic sequence, to complete the system, three other search strategies including single topic, topic co-occurrence, and similar clips are considered in the Retrieval Engine.  
\subsection{Query formulation}
\label{subsec:qForm}
In our framework, users can search for different queries using the proposed query formulation which is as follows \smallskip \\
QUERY DOMAIN \textless Feature space\textgreater \:SECTION \textless Database\textgreater \: SEARCH TYPE \textless Search strategy\textgreater, 
\smallskip \\ 
\noindent where QUERY DOMAIN, SECTION, and SEARCH TYPE are mandatory keywords. \smallskip \\
%
\textbf{\textit{Feature space}} determines visual features that are present in the user query. Since separate models are learned for each visual feature, there is a separate database for each of them. Thus, users can accelerate the search procedure by specifying the features that are present in their query. Users can choose a single or a mixture of feature spaces.  \smallskip \\ 
\textbf{\textit{Database}} term in the query formulation is very important and can improve the search speed. This term determines the part of the dataset that the user is interested in. In the case that the database is divided into the parts related to different locations or different times, this term allows users to narrow down the search space which in turn speeds up the search procedure~\cite{ref_8}.\smallskip \\
\textbf{\textit{Search strategy}} specifies the search scenario from the proposed search strategies. 
\subsection{Query types}
\begin{figure}[!t]
\centering
\includegraphics[width=0.47\textwidth]{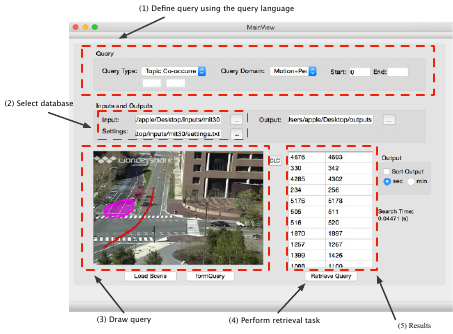}
\caption{Developed software for the retrieval task. It provides a simple way for users to define their queries using the proposed query formulation by drawing paths or specifying regions in the scene.}
\label{fig:app}
\end{figure}

The proposed method supports two query types including \textit{query by sketch} and \textit{query by example}. A software (Fig.~\ref{fig:app}) is developed to provide a simple way for query definition.
\smallskip  \\
\textbf{\textit{Query by sketch:}} In this query type, users can specify the activity they are interested in, by drawing paths or regions in the scene. The developed software provides a straightforward way to do this. For example, Fig.~\ref{fig:app} shows a sample query that indicates co-occurrence of a motion activity and a persistence in the scene. Extracted features from the sketched query are then assigned to the corresponding topics in each model. \smallskip \\
\textbf{\textit{Query by example:}} In this query type, a clip provided by the user represents the query. In this case, the distribution of the input clip forms the user query. Furthermore, the developed software allows users to select a time interval in the database as their query. 
%
%
%
%
\begin{figure}[!t]
\centering
\subfloat[]{\includegraphics[width=1.6in]{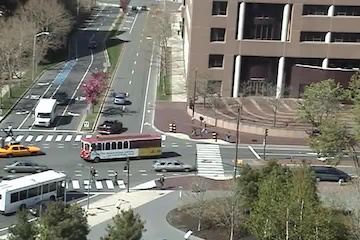}
\label{fig:docBlobDecomp1}} \\
\subfloat[]{\includegraphics[width=1.6in]{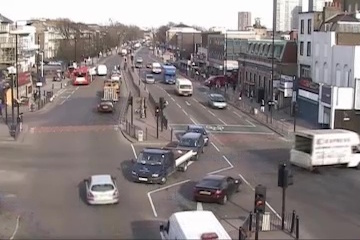}
\label{fig:docBlobDecomp2}}
\hfil
\subfloat[]{\includegraphics[width=1.6in]{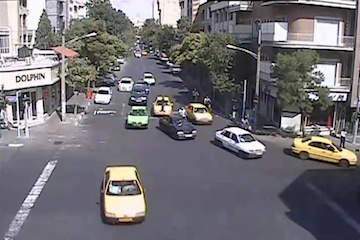}
\label{fig:docBlobDecomp3}}
\caption{Sample frames of tested datasets. (a) MIT Traffic dataset, (b) QMUL Junction dataset, and (c) Tehran Junction dataset.}
\label{fig:datasets}
\end{figure}
%
%
\section{Experimental results}
\label{sec:experiments}  
%
\subsection{Datasets and settings} 
Experiments are carried out on three crowded video sequences recorded by fixed cameras. The first video sequence is MIT Traffic dataset~\cite{ref_23} (Fig.~\ref{fig:datasets}a) including a far-field traffic scene of 92 minutes long. This dataset is recorded at 30 fps with a resolution of $480\times 720$ which is scaled to a frame size of $240\times 360$. The second video sequence is QMUL Junction dataset \cite{ref_23} (Fig.~\ref{fig:datasets}b) of 60 minutes long captured with the frame size of $288\times 360$ at 25 fps. The third tested video sequence is Tehran Junction dataset (Fig.~\ref{fig:datasets}c) of 16 minutes long and resolution of $512\times 720$. This dataset includes activities like stopped vehicles in non-authorized areas. Table \ref{tab:topicNumber} shows the topic number in the primary and secondary models for the presented experiments in the evaluated datasets.  \\ 
%
\begin{table*}[!t]
\renewcommand{\arraystretch}{1.4}
\caption{Topic number for motion and persistence features in the primary and secondary models.}
\label{tab:topicNumber}
\centering
\begin{tabular}{|c||c||c||c||c||c|}
\hline
Video   &  Training  &~Motion topics	& ~~Motion topics   &Persistence topics &Persistence  topics\\
 sequence       	&   data     &  (primary model)  & (secondary model)  & (primary model)   & (secondary model)                  	\\
\hline
MIT Traffic   	&   25 (min)  & 17 	&  49    &15      &30 \\
\hline
QMUL Junction &  20 (min)   &20	&  41    &25         &54\\
\hline 
Tehran Junction &   12 (min)   	&20  &  49     &18     	&48\\
\hline
\end{tabular}
\end{table*}
%
\begin{figure*}[!t]
\centering
\subfloat[]{\includegraphics[width=2.3in]{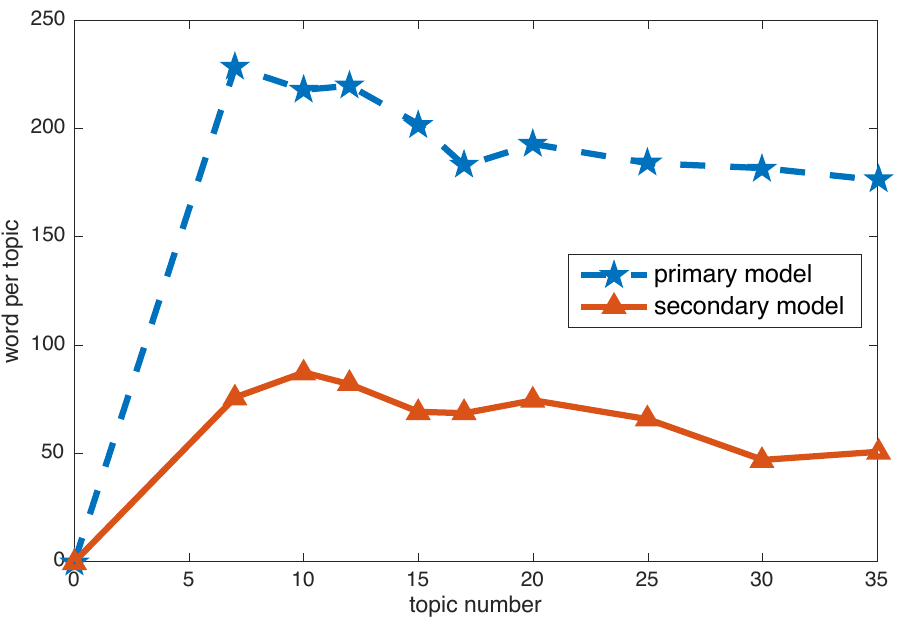}
\label{fig:fig:avgNum1}}
\hfil
\subfloat[]{\includegraphics[width=2.3in]{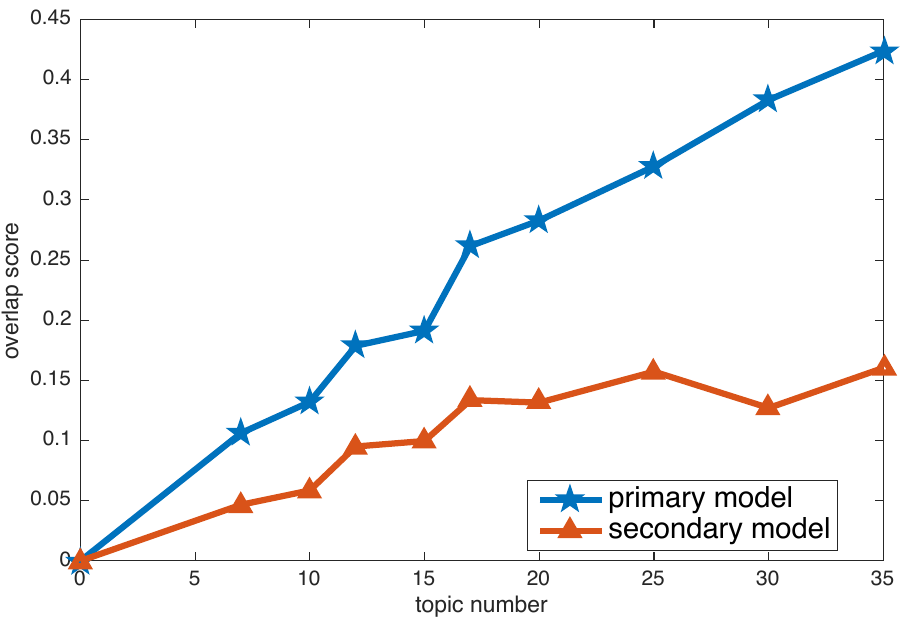}
\label{fig:fig:avgNum2}}
\hfil
\subfloat[]{\includegraphics[width=2.22in]{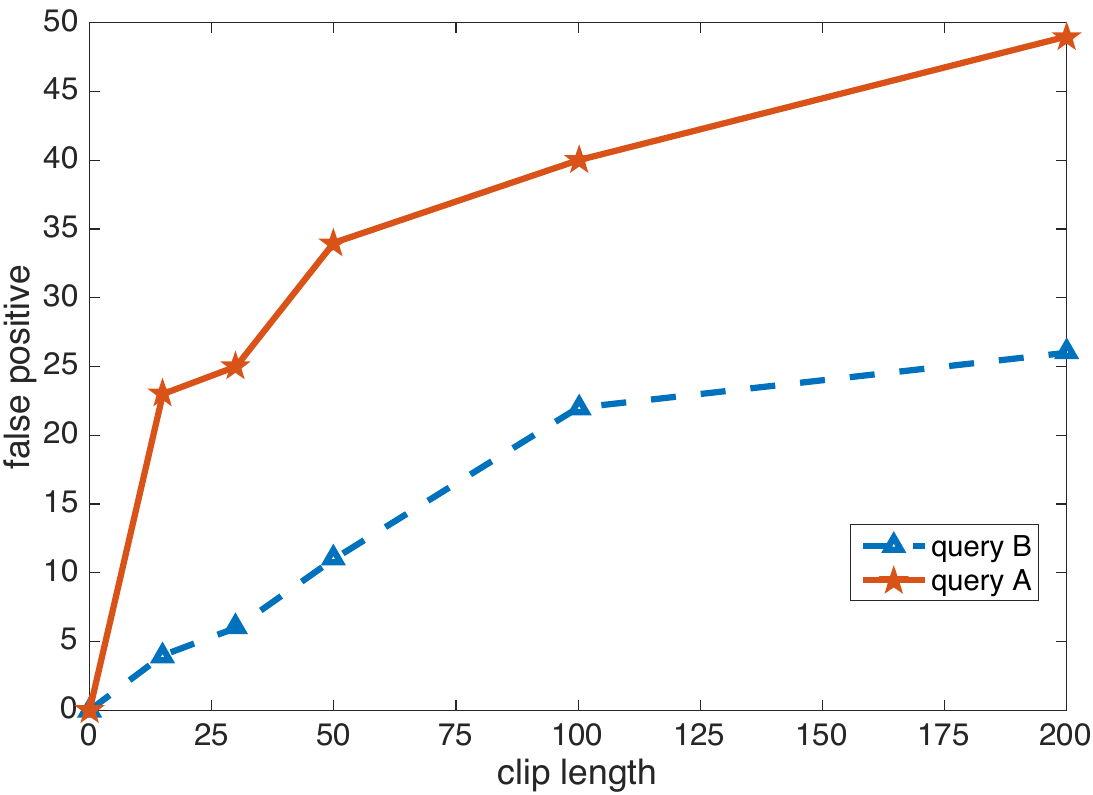}
\label{fig:fig:avgNum3}}
\caption{(a) The average number of words vs the number of topics, (b) overlap score vs the number of topics, (c) clip-length vs the number of false positives.}
\label{fig:avgNum}
\end{figure*}
%
\begin{figure*}[!t]
  \centering
  \includegraphics[width=0.85\textwidth]{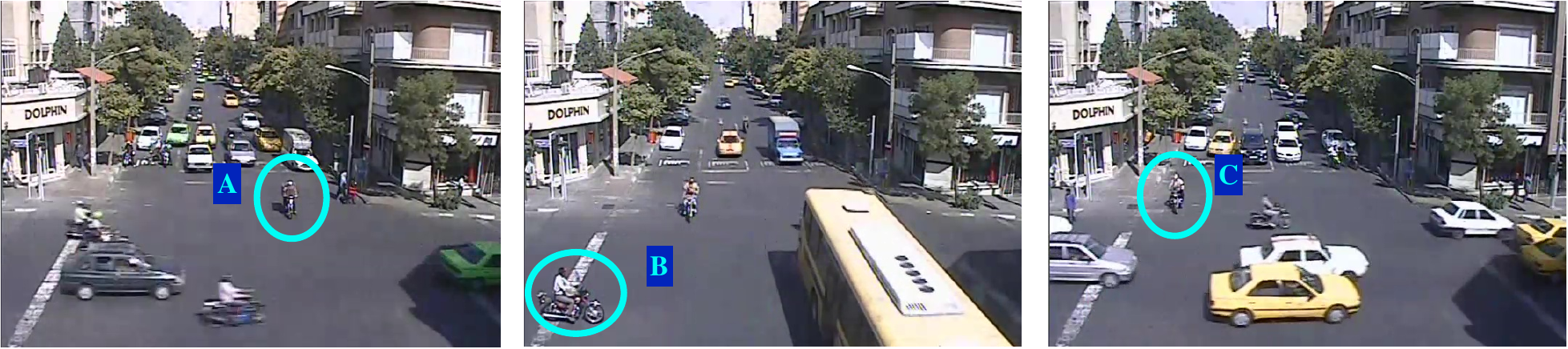}
\caption{Sample persistence queries in Tehran Junction dataset.}
\label{fig:tehranQuery}
\end{figure*}
\textbf{Topic sparsity:} Fig.~\ref{fig:avgNum}a compares the average number of words per topic for different topic numbers in the primary and secondary models. As seen from this figure, topics of the secondary model are sparser that the ones in the primary model. \\ 
\indent \textbf{Topic similarity:} Highly correlated topics can be easily removed in a model with primitive topics, thus, topics in the secondary model have a lower intersection with each other which results in a lower storage. In addition, when the user defines his query by sketch, the user query assigns more easily to uncorrelated topics. A proper evaluation metric for similarity (correlation) between topics is the overlap score \cite{ref_20}. Given two topic blobs $b_i$ and $b_j$, the overlap score is defined as 
%
\begin{align}
S = \frac{|b_i\bigcap b_j|}{|b_i\bigcup b_j|},
\label{eq:overlapScore}
\end{align} 
where $\bigcap$ and $\bigcup$ denote intersection and union operators respectively, and $|.|$ represents the number of pixels constituting the blob of each topic. Fig.~\ref{fig:avgNum}b compares the average overlap score between topics in the primary and secondary models. 
\smallskip \\ 
%
%
\begin{figure*}[!t]
\centering
\subfloat[]{\includegraphics[width=2in]{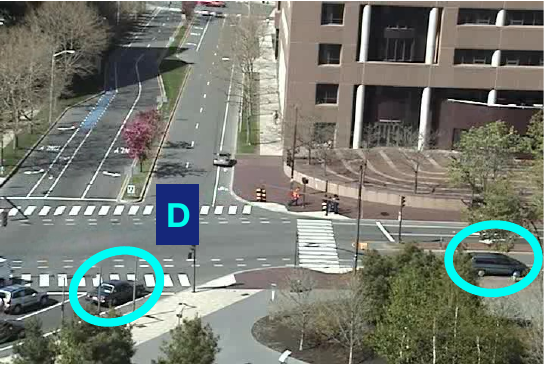}%
\label{fig:coocQuery1}}
\hspace{1em}
\subfloat[]{\includegraphics[width=2in]{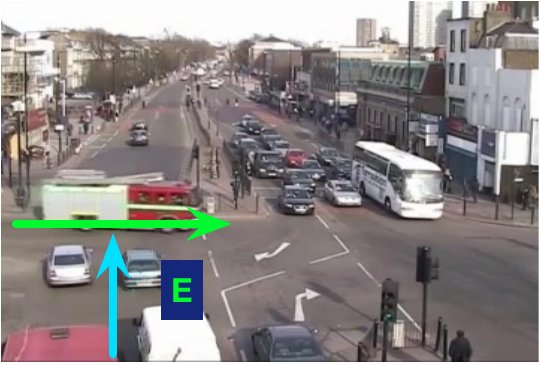}%
\label{fig:coocQuery2}}
\caption{ Co-occurrence of two activities. (a) Persistence co-occurrence in MIT dataset, (b) Traffic interruption by a fire engine in QMUL Junction dataset.}
\label{fig:coocQuery}
\end{figure*}
%
\begin{table*}[!t]
\renewcommand{\arraystretch}{1.2}
\caption{Results for queries searched using single topic and topic co-occurrence strategies.\label{tab:coocSingle}}
\label{tab:coocSingle}
\centering
\begin{tabular}{|c||c||c||c||c||c|}
\hline
~~Query~~  & Video sequence & Ground truth & True positive & False positive & False negative  \\
\hline
A & Tehran Junction & 1 & 1 & 0 & 0 \\
\hline
B & Tehran Junction & 4  & 4 & 0 & 0 \\
\hline
C & Tehran Junction &  5 & 4 &  0 &  1\\
\hline
D & MIT Traffic & 2 & 2 & 0 & 0 \\
\hline
E & QMUL Junction &  1 & 1  & 0 & 0 \\
\hline
\end{tabular}
\end{table*}
\textbf{Clip-length effect:} Clip-length is an important parameter which significantly affects the system performance. In the case of topic sequence search strategy, most false alarm responses occur when two different vehicles traverse the user query. For example, if the user searches for a topic sequence with the length of two, it is likely that two different cars create the topic sequence and cause a false alarm response. Longer clips intensify this issue from two different aspects. First, higher clip-length results in more spatially extended topics. By increasing the clip-length, more displacement is captured in topics, thus, motion topics grow spatially and cover more area in the scene. With spatially broader topics, it is more likely that two different cars hit the topics included in the user query. Secondly, regardless of topics expansion, higher clip-length provides a longer time window for the second car to complete the sequence and cause a false positive response. Fig.~\ref{fig:clipLengthChart} shows the search result for queries (F) and (G) illustrated in Fig.~\ref{fig:mitQueries} for databases created with different clip-lengths. As seen from Fig.~\ref{fig:avgNum}c, the number of false positive responses increases in higher clip-lengths. 
\subsection{Evaluated queries}
\label{sub_evaluatedQueries}
\begin{figure}[!t]
  \centering
  \includegraphics[width=0.32\textwidth]{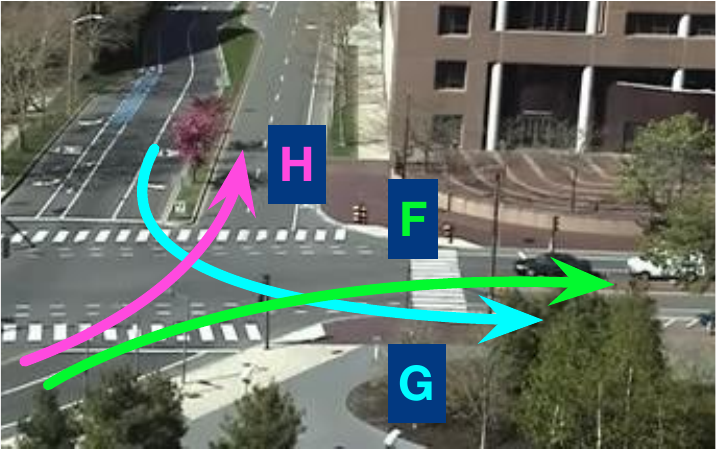}
\caption{Queries defined as a topic sequence in MIT Traffic dataset.}
\label{fig:mitQueries}
\end{figure}

\noindent \textbf{Single topic:} Querying a single motion topic is the same as querying a topic sequence with the sequence length of one. However, in other feature spaces, our desired activity can be specified by a single topic. For example, stopped vehicles in non-authorized areas can be found by querying the related persistence topic in the persistence database. Consecutive clips which each of them includes the queried topic, merge together to form the full response. Fig.~\ref{fig:tehranQuery} shows persistence queries in no waiting areas in Tehran Junction dataset, and the results of these queries are demonstrated in Table~\ref{tab:coocSingle}.  \\ 
\noindent \textbf{Topic co-occurrence:} Co-occurrence of two activities is another search strategy which is considered in our framework. In this strategy, the simultaneous occurrence of two actions in a single clip is considered. 
Fig.~\ref{fig:coocQuery} shows two queries defined using the topic co-occurrence strategy. Fig.~\ref{fig:coocQuery1} shows a query consisting of persistence in two different location in MIT Traffic dataset, and a traffic interruption by a fire engine in QMUL Junction dataset is illustrated in Fig.~\ref{fig:coocQuery2}. Table~?? illustrates the search results for these queries (D and E). Consecutive clips which each of them includes the queried topics, merge together to form the full response.  \\ 
%
%
\noindent \textbf{Topic sequence:} A primitive topic can not be used individually to represent a complicated activity. Thus, in the proposed method, a complicated activity is defined as a topic sequence.
%
Fig.~\ref{fig:mitQueries} shows three queries in MIT Traffic dataset, and Table~\ref{tab:topicSeq} presents the search results of our method for these queries. The results of the method based on HDP topic model \cite{ref_7} which is evaluated in \cite{ref_2}, and the results of the method based on low-level features \cite{ref_2} are also included.
%
%
\begin{figure}[!t]
  \centering
  \includegraphics[width=0.32\textwidth]{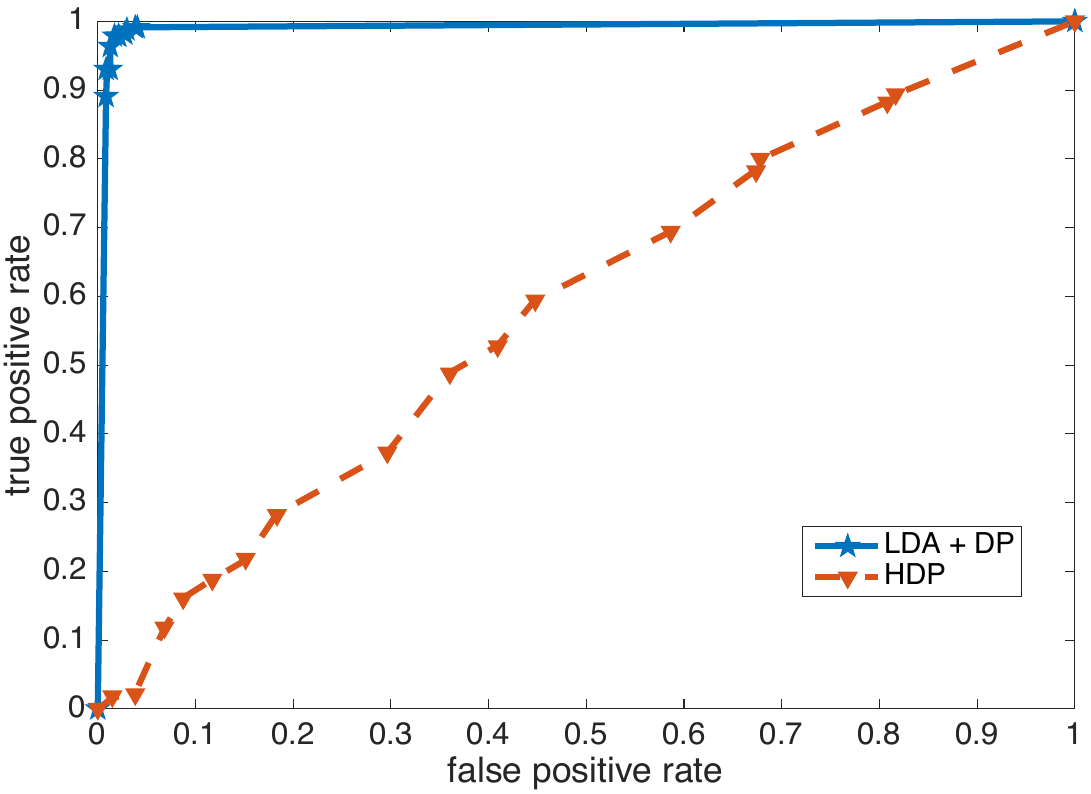}
\caption{ROC curves for query G illustrated in Fig.~\ref{fig:mitQueries}.}
\label{fig:roc}
\end{figure}
%
\begin{table*}[!t]
\renewcommand{\arraystretch}{1.3}
\caption{Results for queries searched using topic sequence search strategy.}
\label{tab:topicSeq}
\centering
\begin{tabular}{|c||c||c||c||c||c||c||c|}
\hline
Query  & Ground  & This work &  HDP \cite{ref_7} & LL features \cite{ref_2}  & This work & HDP \cite{ref_7} & LL features\cite{ref_2} \\
          	&   truth                  	& (true positives) 	&  	(true positives)       	&  (true positives)   & (false alarms) 	& (false alarms)                 	& (false alarms) \\
\hline
F  & 148 & 121 & 54 & 135 & 23 & 118 &13\\
\hline
G & 66  & 63 & 6 & 61 & 4 & 58 &  5\\
\hline
H & 170 & 153 & --- & --- & 4 &--- & ---\\
\hline
\end{tabular}
\end{table*}
%
\begin{figure*}[!t]
  \centering
  \includegraphics[width=0.9\textwidth]{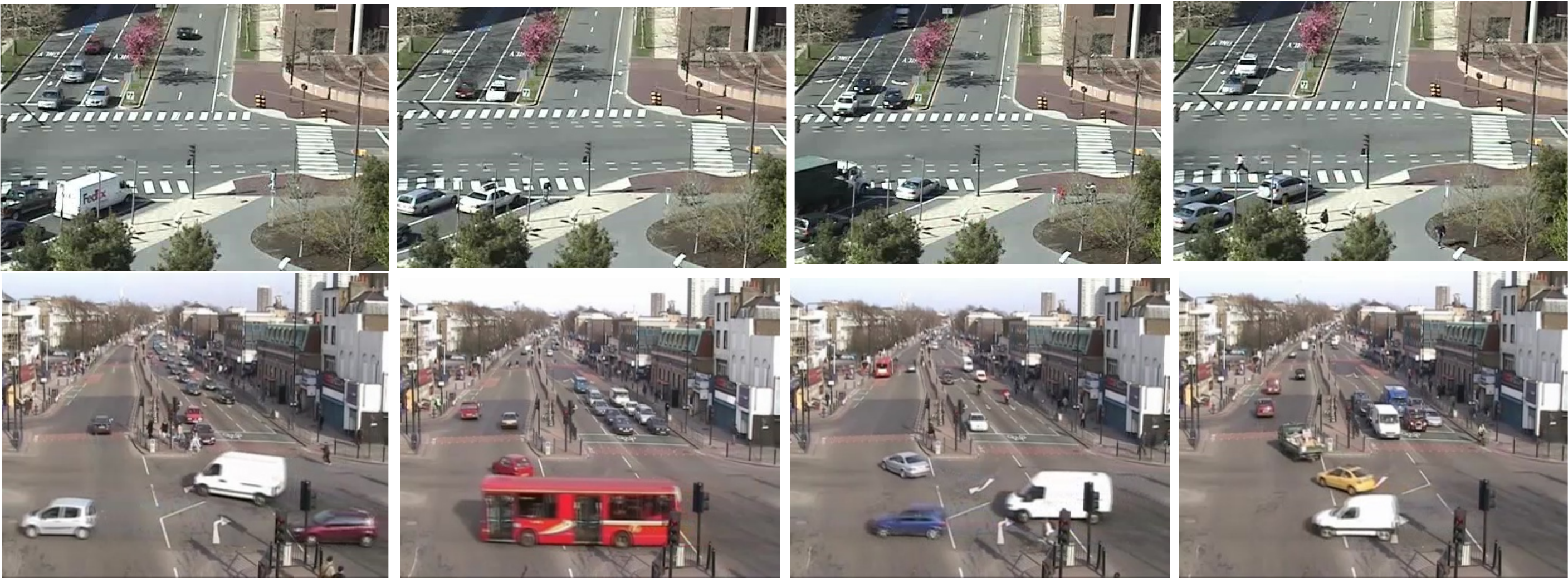}
\caption{Similar scenes. (First row): MIT Traffic dataset, (Second row): QMUL Junction dataset. }
\label{fig:simClipsSample}
\end{figure*}
Table~\ref{tab:topicSeq} demonstrates a substantial improvement over topic models performance in the retrieval task. For activity (F), this improvement accounts for $124\%$ increase in true positive and $80\%$ decrease in false positive responses. For activity (G), there is a $950\%$ increase in true positive and $93\%$ decrease in false positive responses. The area under ROC (AUROC) value for activity (G) increases from 0.63 in \cite{ref_7} to 0.89 in our work. The AUROC for activities (F) and (H) in our method are 0.91 and 0.80 respectively. Fig. \ref{fig:roc} shows ROC curves for activity (G) in our method and \cite{ref_7}.  \\
%
\noindent \textbf{Similar clips:} Similar clips to a given clip are found using topic distribution comparison. The topic distribution of the queried clip $q$ is matched with topic distribution of the database clips $p_c$ using the Hellinger distance: \\
\begin{align}
d_{H}=\frac{1}{\sqrt{2}}\sqrt{\sum_{i=1}^{N}(\sqrt{p_{ci}}-\sqrt{q_{i}})^2}.
\label{eq:hell}
\end{align} 
After computing the Hellinger distance, the results are sorted ascending. 
In the case that the queried clip has a longer length than those in the database, the queried clip is broken down into multiple clips with the same length as clips in the dataset. For example, in our experiments, MIT Traffic dataset is divided into clips with one-second length. Thus, in order to find similar scenes to a three-seconds clip, the queried clip is divided into three clips each of them one second long, and the resulted distributions are convoluted over the database clips. Fig.~\ref{fig:simClipsSample} shows sample frames of two queried scenes in MIT Traffic and QMUL Junction datasets (first column). Frame samples of the first three clips which are most similar to the queried scenes in the corresponding databases are also shown in this figure (columns 2 to 4). 
%
\subsection{Database storage and search speed}
\begin{figure}[!b]
\centering
\includegraphics[width=0.35\textwidth]{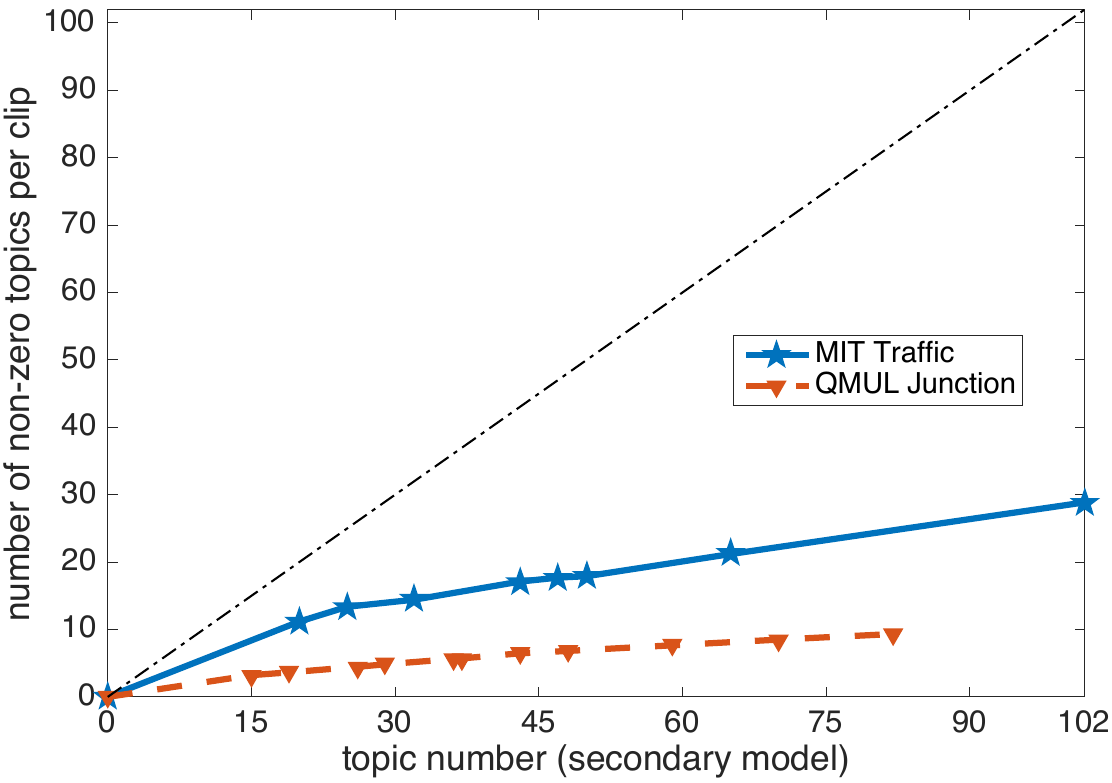}
\caption{The average number of non-zero topics per clip.}
\label{fig:topicPerClip}
\end{figure}

\indent Indexing video with high-level primitive topics, not only provides a fine description of the input video, but it also results in a lightweight database in comparison to the methods based on low-level features like \cite{ref_2}. Table~\ref{tab:storage} compares database size for MIT Traffic dataset in our work and  \cite{ref_2}. In a database which includes clip distributions over all of the learned topics, the database size grows linearly with the size of the input video, regardless of its content. Since clips with not significant content have sparse distributions, the database size can be shrunk, if we discard elements which are lower than a specified threshold. In other words, in each clip, we store topic index and topic portion (\textit{topic-idx,topic-value}) for the ones with significant values. By doing so, the database size is only dependent on the content of the scene's foreground. Fig.~\ref{fig:topicPerClip} demonstrates the average number of non-zero elements in MIT Traffic and QMUL Junction datasets. As can be seen from this figure, the database size grows sub-linearly with the topic number in the secondary model, and it means that clips with not significant content have a sparse distribution over topics of the secondary model. Table~\ref{tab:storage} compares the database size in our work and \cite{ref_2}. 
%
\begin{table}[!b]
\renewcommand{\arraystretch}{1.43}
\caption{Database size comparison. (MIT traffic dataset)}
\label{tab:storage}
\centering
\begin{tabular}{|c||c|}
\hline
Method & Database Size\\
\hline
Low-level features \cite{ref_2} & 42 MB \\
\hline
This work  & 1.04 MB \\
\hline
This work (non-zeros)  & 0.56 MB \\
\hline
\end{tabular}
\end{table}
%
%
\begin{table}[!b]
\renewcommand{\arraystretch}{1.4}
\caption{Search time for queries (F), (G) and (H) shown in Fig. \ref{fig:mitQueries}.\label{tab:searchSpeed}}
\label{tab:storage}
\centering
\begin{tabular}{|c||c||c||c|}
\hline
Query & F & G & H\\
\hline
Low-level features \cite{ref_2} & 0.5 (s) & 0.4 (s) & -- \\
\hline
This work  & 0.029 (s) & 0.026 (s) &  0.016 (s) \\
\hline
\end{tabular}
\end{table}
%
A lightweight database results in a speed~up in the search procedure. Table~\ref{tab:searchSpeed} shows the search time in our work compared to the method in \cite{ref_2}. As can be seen from the table, the search procedure in our method is considerably fast. 
\subsection{Accelerating feature extraction using GPU}
\label{sub_gpuSpeedsUp}
\begin{figure}[!t]
\centering
\includegraphics[width=0.2\textwidth]{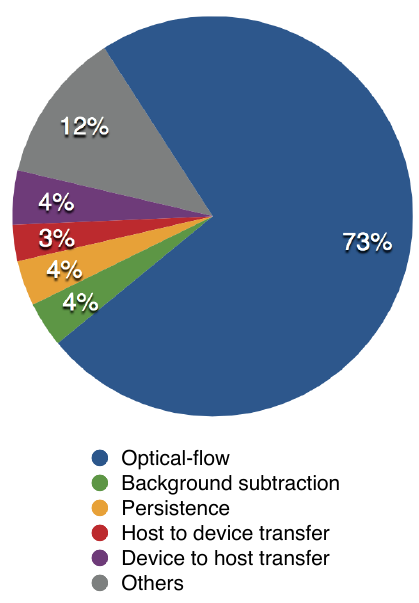}
\caption{Computation time proportion for each step of feature extraction step performed on GPU.}
\label{fig:feChart}
\end{figure}

\indent The feature extraction speed for MIT Traffic dataset is about 24 frames per second at the frame size of $240\times 360$. Table~\ref{tab:compTimeFE} shows the computation time for each step of feature extraction and compares the execution time of each step on CPU and the parallel version on GPU.  As can be seen from Table~\ref{tab:compTimeFE}, feature extraction is about 4.5 times faster when it executes on GPU. Host to device and device to host transfers indicate the time of data transfers from CPU to GPU and from GPU to CPU respectively. Fig.~\ref{fig:feChart} shows that the computation of optical flow is the most time-consuming step of feature extraction step.
%
%
%
\begin{table}[!t]
\renewcommand{\arraystretch}{1.43}
\caption{Feature extraction computation time. (unit: ms)}
\label{tab:compTimeFE}
\centering
\begin{tabular}{|c||c||c|}
\hline
Feature extraction step & CPU & GPU  \\
\hline
Optical-flow & 173 &  30  \\
\hline
Background subtraction & 5  &  1.5  \\
\hline
Persistence & 5  &  1.5  \\
\hline
Host to device transfer & --  & 1.2 \\
\hline
Device to host transfer &  -- &  1.8 \\
\hline
Other &  5  & 5 \\
\hline
Total &  183  & 41 \\
\hline
\end{tabular}
\end{table}

\section{Conclusion}
\label{sec:conclusion}
In this paper, we addressed the ambiguity issue in topic models. For this purpose, several techniques are used to create a model which describes the scene without any ambiguity. Learning separate models for each visual feature prevents activities from different feature spaces to mix together in a single topic. In addition, a pre-processing step and several topic processing techniques are used to create a model in which topics present primitive actions each of them comprising a single blob with movement in a single direction (in the case of motion feature). In the proposed method, four search strategies including topic sequence, single topic, topic co-occurrence, and similar clips are proposed. A variety of queries can be defined using the proposed query formulation. The proposed formulation provides a straightforward way for users to interact with the system, and it can also speed up the search procedure by narrowing down the search space. In the proposed method, a complicated activity can be defined as a topic sequence that is searched using dynamic programming. Defining activities as a sequence of primitive topics brings a substantial performance improvement in the retrieval task compared to the other methods based on topic models. Furthermore, a lightweight database, not only occupies much fewer storage in our method, but it also results in a significant speed up in the search procedure compared to the methods which are based on low-level features. The feature extraction step which is computationally demanding performs in real time by leveraging the computing power of GPU. 
%

%



%
%

\ifCLASSOPTIONcaptionsoff
  \newpage
\fi

\end{document}